\documentclass[]{spie}  
\usepackage[]{graphicx}

\usepackage{amsmath}
\usepackage{amssymb}
\usepackage[square,comma,numbers]{natbib} 
\usepackage{color,soul} 
\usepackage{hyperref}

\usepackage{float} 

\usepackage{booktabs} 
\usepackage{rotating}
\usepackage{multirow}

\title{Computer-assisted polyp matching between optical colonoscopy and CT colonography: a phantom study} 

\author{Holger R. Roth\supit{a}, Thomas E. Hampshire\supit{a}, Emma Helbren\supit{b}, Mingxing Hu\supit{a}, Roser Vega\supit{c}, Steve Halligan\supit{b} and David J. Hawkes\supit{a}
\skiplinehalf
\supit{a}Centre for Medical Image Computing, University College London, UK \\
\supit{b}Centre for Medical Imaging, University College London, London, UK \\
\supit{c}Gastrointestinal Services, University College Hospital, London, UK
}

\authorinfo{Further author information: Send correspondence to Holger R. Roth: E-mail: h.roth@ucl.ac.uk}

\begin{document} 
  \maketitle 

\begin{abstract}
Potentially precancerous polyps detected with CT colonography (CTC) need to be removed subsequently, using an optical colonoscope (OC). Due to large colonic deformations induced by the colonoscope, even very experienced colonoscopists find it difficult to pinpoint the exact location of the colonoscope tip in relation to polyps reported on CTC. This can cause unduly prolonged OC examinations that are stressful for the patient, colonoscopist and supporting staff.

We developed a method, based on monocular 3D reconstruction from OC images, that automatically matches polyps observed in OC with polyps reported on prior CTC. A matching cost is computed, using rigid point-based registration between surface point clouds extracted from both modalities. A 3D printed and painted phantom of a 25 cm long transverse colon segment was used to validate the method on two medium sized polyps. Results indicate that the matching cost is smaller at the correct corresponding polyp between OC and CTC: the value is 3.9 times higher at the incorrect polyp, comparing the correct match between polyps to the incorrect match. Furthermore, we evaluate the matching of the reconstructed polyp from OC with other colonic endoluminal surface structures such as haustral folds and show that there is a minimum at the correct polyp from CTC. 

Automated matching between polyps observed at OC and prior CTC would facilitate the biopsy or removal of true-positive pathology or exclusion of false-positive CTC findings, and would reduce colonoscopy false-negative (missed) polyps. Ultimately, such a method might reduce healthcare costs, patient inconvenience and discomfort.
\end{abstract}


\keywords{optical colonoscopy, virtual colonoscopy, CT colonography, 3D reconstruction, registration, computer-aided diagnosis and interventions}

\section{Introduction}
Detection of potentially precancerous polyps and cancers using CT colonography (CTC) is only the first step of patient treatment. Once a suspected polyp has been detected using CTC, the patient must undergo subsequent optical colonoscopy (OC) to enable polypectomy -- surgical removal of the polyp. However, CTC can suffer from false-positive polyps reported by radiologists during its interpretation and such polyps will not be found during subsequent OC. Therefore a polyp seen at CTC may not be seen at subsequent colonoscopy for either of two reasons: 1.) a polyp is not actually present, i.e. a false-positive CTC diagnosis; 2.) a polyp is present, but the colonoscopist cannot find it. 

In present clinical practice, the colonoscopist does not know which of these two circumstances apply when a polyp reported on CTC cannot be found during subsequent OC. Previous studies of back-to-back OC indicate that the miss-rates for 10 mm adenomas or larger is approximately 6\% \cite{rex1997colonoscopic}. Due to deformations induced by the colonoscope, variable colon lengths and differences in haustration, even very experienced colonoscopists find it difficult to pinpoint the precise location of the endoscope tip during the examination. This makes it hard to relate the colonoscopic video to the radiological findings from CTC. This problem can cause unduly prolonged OC examinations that are stressful for the patient, colonoscopist and supporting staff.

In this study, we performed a phantom study that investigated the feasibility of automatically matching polyps and other abnormalities perceived during OC with polyps reported on prior CTC.

\subsection{Related work}
\citet{hong2011colon} propose an algorithm that detects the edges of haustral folds using a single OC image. They extract rings of fold contours by fitting Bezier curves. These contours are then used to reconstruct a virtual 3D representation of a colonic segment, using a reverse-projecting of the 2D fold contours into 3D space, as described in their previous work \citep{hong20093d}. They propose to use this method for estimating how much area of the colonic surface mucosa a colonoscopist has observed during OC. However, their method relies on the assumption that the diameter of fold contours and the colon does not change in the segment observed in a single OC image. Therefore, a simple reverse-projection of the fold contours will give a reasonable estimate of the fold position in 3D -- an assumption certainly not true for the entirety of the colon observed in colonoscopy. 

\citet{liu2013optical} describe a purely computer-vision based method for tracking the motion of an colonoscope in relation to virtual colonoscopy. After manual initialization between the virtual view in VC and the current location of the colonoscope tip, they estimate the motion of the colonoscope using an optical-flow-based method that ignores non-informative (blurry) image frames \citep{liu2008optic-flow}. Their exclusive use of vision-based information avoids the usage of any external tracking sensors such as electromagnetic tracking. However, \citet{liu2013optical} describe certain limitations of their current method: it relies on good manual initialization between the first OC frame to be tracked and the VC view. A good initialization is mainly achieved by using colonic regions with clear landmarks such as polyps that need to be linked by experts in both OC and VC. Otherwise, their optical flow algorithm relies on the fold structures of the colon which can be misleading due to is tubular structure and repeating fold shapes \citep{liu2008optic-flow}. Furthermore, the colon is highly flexible in the sigmoid and transverse colon and here, \citet{liu2013optical} describe problems when tracking the colonoscope due the large deformations and ``sharp turns'' of the colonoscope when it passes these regions. A continuous tracking of the colonoscope using optical flow proved very challenging in these regions of the colon and therefore several re-initializations of the method would need to take place. Therefore, more automated methods of initialization between the optical and colonoscopic images would need to be implemented in order to make this method work in clinical practice. 

The work presented here was motivated by this issue of providing the means of automated initialization between OC and VC images. Previous work in optical colonoscopy exists on camera motion estimation (potentially usable for 3D reconstruction) \citep{sargent2009feature,sargent2011image} and sparse 3D reconstruction \citep{koppel2007toward, gu2007computer, chen2009uniscale}, but no attempt has been made to use that information in order to match with VC. Furthermore, it is likely that a sparse 3D reconstruction will not deliver the required detail to accurately match to VC data.

\section{Methods}
\begin{figure}[H]
\centering	\resizebox{0.8\textwidth}{!}{\includegraphics{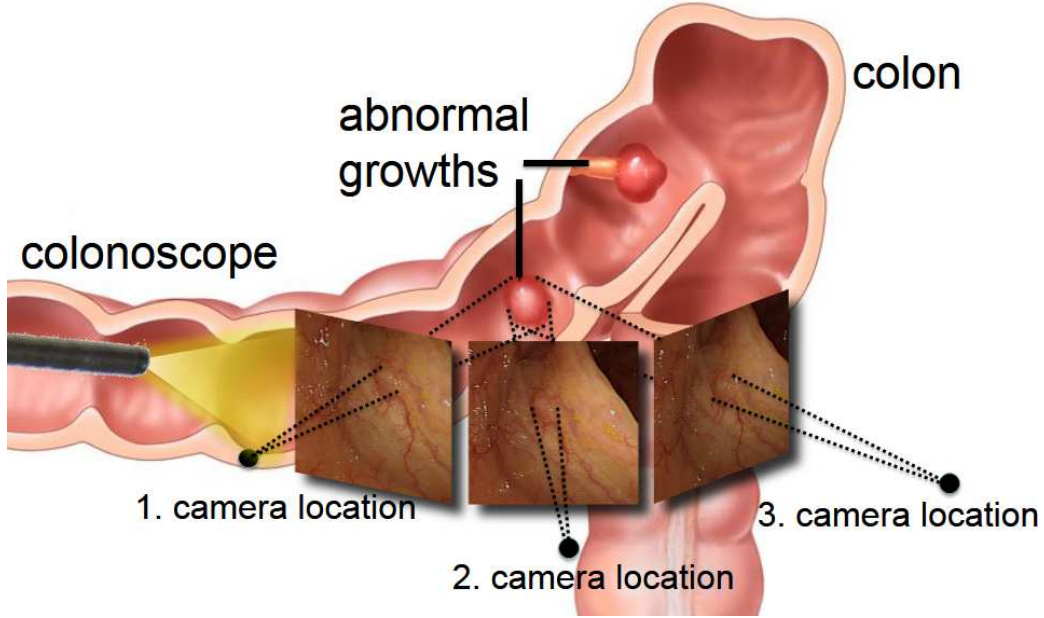}}
	\caption{3D reconstruction of an abnormal polyp from three colonoscopy images.}					
	\label{fig:reconstruction_principle}
\end{figure}

Recent advances in computer vision (CV) enable us to compute dense 3D structural information from 2D camera images. Structure-from-motion approaches for 3D reconstruction use the perspective shift between camera images observing the same three-dimensional scene. After pre-calibration of the camera, different viewpoints of the same structure allow the triangulation of 3D points of the corresponding projections of these points in the camera images if the relative translation and rotation of the cameras are known \cite{hartley2000multiple}. Figure \ref{fig:reconstruction_principle} illustrates the reconstruction process of the 3D shape of a polyp using three colonoscopy camera images, showing the same polyp from different viewing angles. Automatic matching between the different images allows the 3D reconstruction of the polyp's shape and the camera's current location with respect to the polyp. The real shape of the polyp can then be reconstructed by triangulation between its surface points in 3D space and the camera locations. 
 
\subsection{Dense 3D reconstruction}
\begin{figure}[htb!]
	\begin{center}
		\resizebox{0.5\textwidth}{!}{\includegraphics{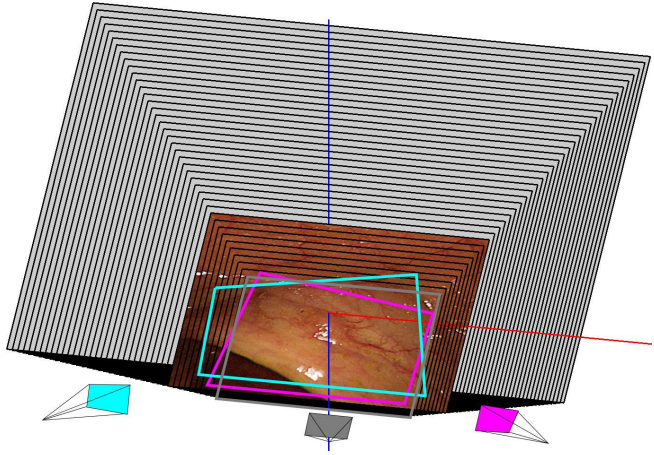}}
		\caption{Plane-sweep approach used for multi-view 3D reconstruction. The 3D space is discretized into several parallel planes. This method allows us to measure the dissimilarity of back-projected points (pixels) at these planes \citep{collins1996space,yang2005versatile}.}
		\label{fig:plane_sweep}
	\end{center}
\end{figure}

We used a dense 3D reconstruction algorithm (specifically developed for narrow-baseline applications) based on an open-source implementation by Paalanen et al. \cite{paalanen2010narrow} that uses the plane-sweep approach proposed by Collins \cite{collins1996space} (Figure \ref{fig:plane_sweep} illustrates the principles of the plane-sweep approach). This approach has the advantage that it is parallelizable and can therefore be used to reconstruct surfaces quickly from multiple camera views using commodity graphics hardware as introduced by Yang and Pollefeys \cite{yang2005versatile}. The plane-sweep approach assumes a plane in 3D space onto which OC images can be back-projected. Corresponding points from the same objects in the colonoscopic scene can then be back-projected onto the same plane in 3D. This can be measured by defining a matching cost $\zeta(x,y)$ at each pixel $(x,y)$ that measures the dissimilarity of the back-projected points. This cost is computed for each pixel in the images at several depths as in 
\begin{equation}
	\zeta(x,y) = \frac{1}{N_\mathrm{V}}\left(\sum^{N_\mathrm{V}}_{i=1}\boldsymbol{I_i}(u_i,v_i)^\mathrm{T} \boldsymbol{I_i}(u_i,v_i)\right) - \boldsymbol{\textsl{s}}^\mathrm{T}\boldsymbol{\textsl{s}},
\end{equation}
where
\begin{equation}
	 \boldsymbol{\textsl{s}} \equiv \frac{1}{N_\mathrm{V}}\sum^{N_\mathrm{V}}_{i=1}\boldsymbol{I_i}(u_i,v_i),
\end{equation}
$N_V$ is the number of views and $I_i$ are the gray images at each view $i$ as in \cite{paalanen2010narrow}.

The resulting reconstructed pixel depths are the ones minimizing the dissimilarity cost across all planes \cite{paalanen2010narrow}. We ignore specular reflections when tracking the motion of the OC camera by detecting areas of high intensity in the OC images using thresholding.

\subsection{Point set registration}
After dense 3D reconstruction of the endoscopic scene, we use the rigid version of Coherent Point Drift (CPD). CPD is a point set registration algorithm, that allows us to match the polyp point clouds reconstructed from OC to the targeted polyp surfaces as extracted from CTC. CPD uses the L2 norm between the two sets of points and probabilistic methods with Gaussian mixture models \cite{myronenko2010point} in order to minimize a squared distance between both point sets: 
\begin{equation}
	\sigma^2 = \frac{1}{DNM}\sum_{n=1}^N\sum_{m=1}^M{\left\|x_n - y_n\right\|}^2,	
\end{equation}
where $x$ and $y$ are the target and source point sets and $N$, $M$ are the number of target and source points, respectively. D is the dimensionality of the points -- 3 in this case.

\section{Experimental evaluation}
\subsection{Colon phantom generation}
\begin{figure}[htb!]
	\begin{center}	\resizebox{0.6\textwidth}{!}{\includegraphics{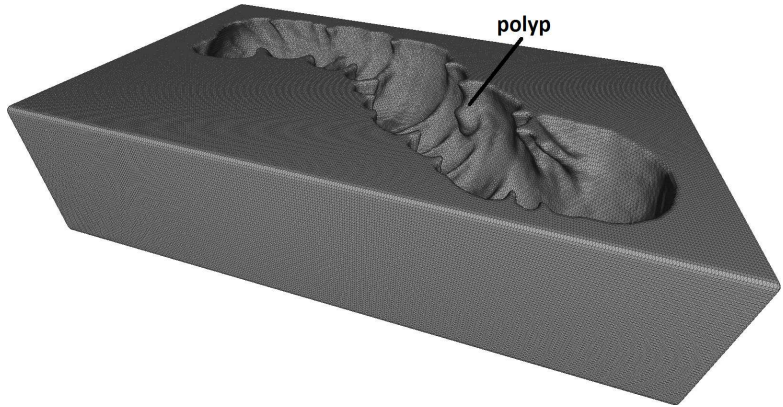}}
		\caption{Underlying 3D stereolithography model for 3D-printing a colon phantom.}
		\label{fig:colon_phantom_model}
	\end{center}
\end{figure}

\begin{figure}[htb!]
	\begin{center}	\resizebox{0.6\textwidth}{!}{\includegraphics{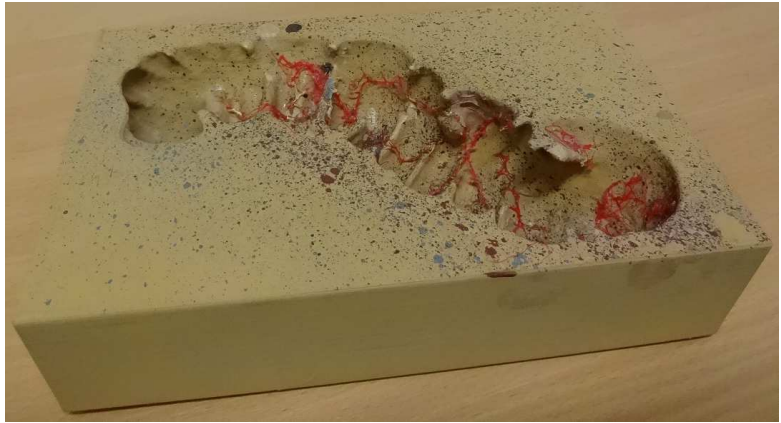}}
		\caption{We painted a colon phantom after 3D-printing in order to mimic the visual features of colonic mucosa.}
		\label{fig:colon_phantom_painted}
	\end{center}
\end{figure}

A colon phantom was produced using 3D printing (MakerBot) from a stereolithography model generated from real CTC data of a 25 cm long section of transverse colon with two polyps (8 mm and 15 mm diameter). The colon was printed in two halves such that blood vessels and colon mucosa could be painted on the inside of the model in order to mimic the visual features of colonic mucosa. Figure \ref{fig:colon_phantom_model} shows the underlying 3D stereolithography model that was used for printing one of the two sides of the colon. After painting (see Fig. \ref{fig:colon_phantom_painted}), both halves were assembled to form a closed colon model into which an endoscope could be inserted. 

\newpage
\subsection{Endoscopic video acquisition and 3D reconstruction}
\begin{figure}[H]
\centering	\resizebox{1.0\textwidth}{!}{\includegraphics{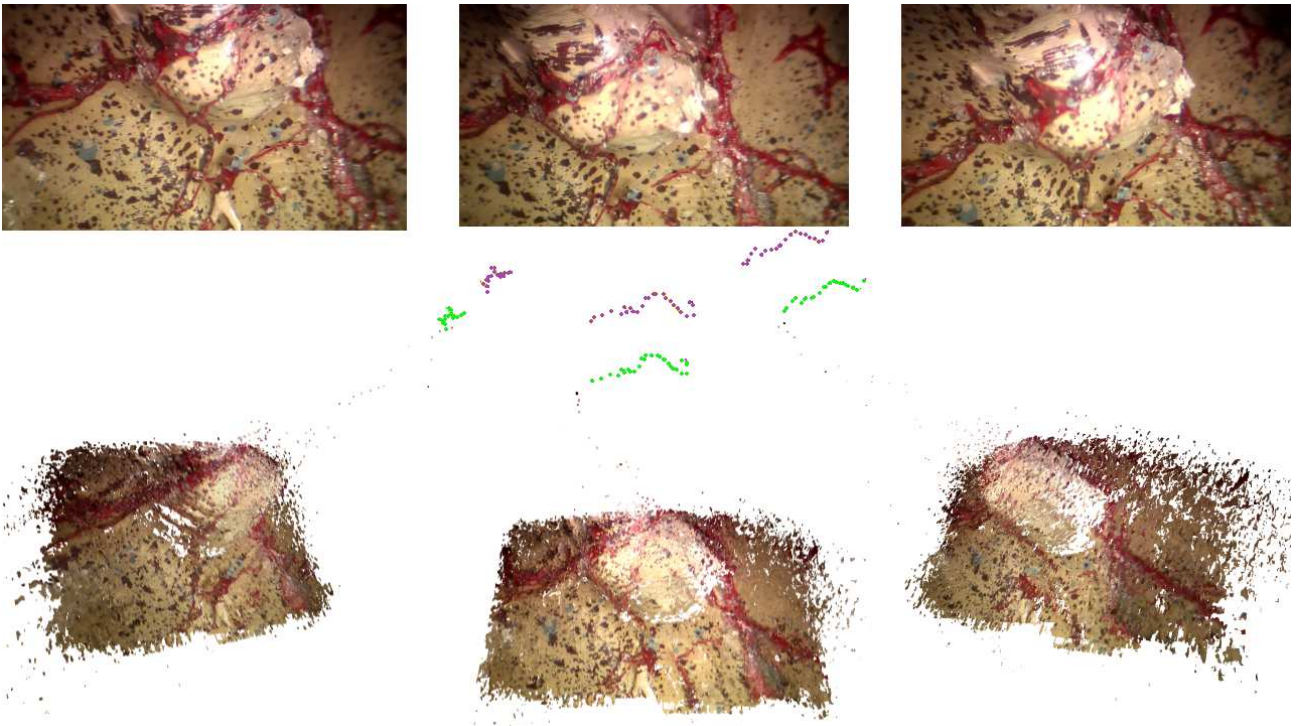}}
	\caption{3D reconstruction of a rigid colon phantom surface at a 15 mm polyp in the transverse colon. Top: three example images of the polyp from optical colonoscopy (OC). Bottom: 3D reconstructed point cloud from three different views. The tracked OC camera positions are plotted using magenta dots (green dots indicate camera viewing direction)}					
	\label{fig:slamrecon}
\end{figure}

We used one video channel (as with a standard monocular colonoscope) of a \textit{VIKING} laparoscope with 1920x539 resolution and recorded a short sweep motion of the camera around the 15 mm polyp as the input sequence to the 3D reconstruction algorithm. For 3D reconstruction in OC, we used only the upper field of the interlaced video and all video frames were resized to half size in order to achieve real-time processing speed (at 15 fps using up to 32 views for 3D reconstruction). Figure \ref{fig:slamrecon} shows the 3D reconstruction of a colonic surface with the 15 mm polyp acquired using the rigid phantom.  

\subsection{Matching with targeted polyp}
After 3D reconstruction of the 15 mm polyp observed at OC, the reconstructed point cloud can be registered to polyp surface point clouds extracted from CTC, using a segmentation of the colonic lumen \cite{roth2011registration}. Running the rigid CPD registration against all polyp locations from CTC could potentially provide a measurement that indicates which CTC polyp corresponds to the 3D reconstructed point cloud around the polyp as observed in OC. Table \ref{tab:registration_errors} shows the result of the $\sigma$ measurement (square root of $\sigma^2$) for the 8 mm and 15 mm CTC polyp and OC point cloud of the 15 mm polyp (as shown in Fig. \ref{fig:slamrecon}). One can see that $\sigma$ value is 3.9 times higher at the incorrect polyp, comparing the correct match between polyps to the incorrect match. 

\begin{table}[htbp]
  \centering
	\caption{Rigid Coherent Point Drift (CPD) registration of a 3D reconstructed point cloud from 	optical colonoscopy (OC), with two polyp surface point clouds extracted from CT colonography (CTC). The tables show the distance measure $\sigma$ that could provide an indication of how well reconstructed OC points and the CTC polyp point clouds are aligned. Absolute $\sigma$ values are shown on the left, while the right table shows the values normalized with $\sigma$ value at the correct polyp match (colored in green). It shows that at the incorrect polyp, the value of $\sigma$ is 3.9 times larger.}
   \begin{tabular}{c}
		\includegraphics[width=0.75\textwidth]{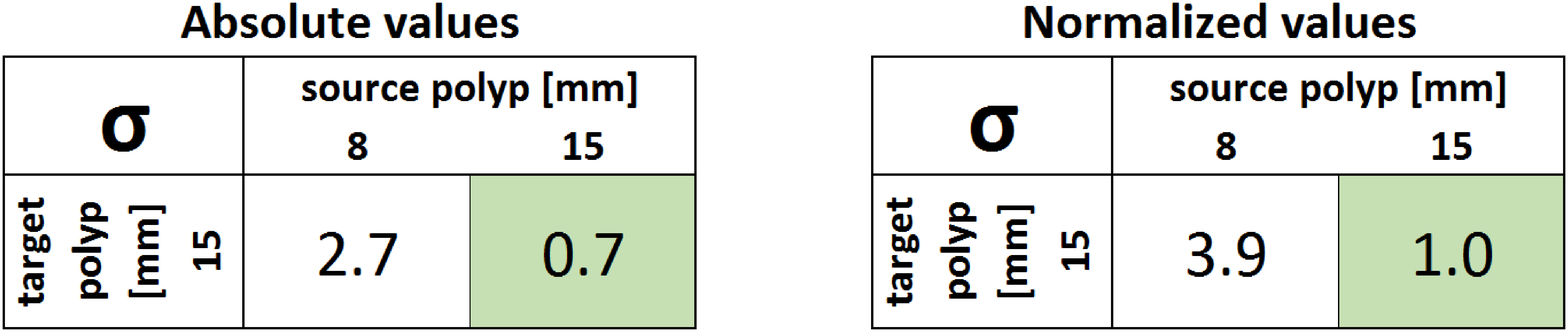}
    \end{tabular}%
  \label{tab:registration_errors}%
\end{table}%

\subsection{Matching with other colonic surface structures}
\begin{figure}[H]
\centering	\resizebox{1.0\textwidth}{!}{\includegraphics{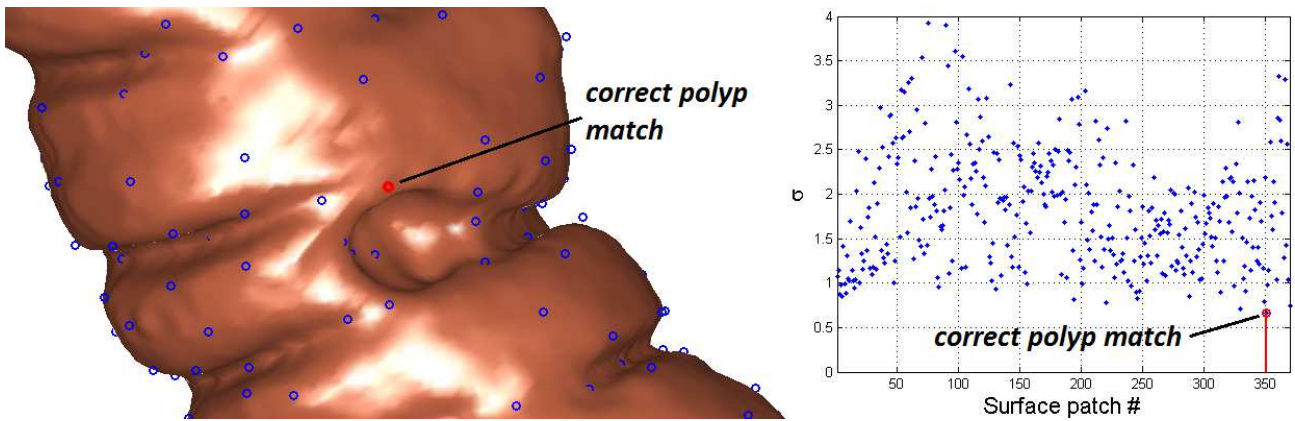}}
	\caption{Comparison of $\sigma$ value between 3D reconstructed point cloud from optical colonoscopy (OC) and other colon surface structures such as haustral folds and the targeted 15 mm polyp on a short section of transverse colon, extracted from CT colonography. Polyp candidates are shown in blue. The correctly matched polyp is shown in red (the 3D reconstruction from OC views the polyp from the side as seen in Fig. \ref{fig:slamrecon}).}					
	\label{fig:result}
\end{figure}

Figure \ref{fig:result} shows a comparison of $\sigma$ value when registering the 3D reconstructed point cloud (15 mm polyp) from OC rigidly with other colonic surface structures, such as haustral folds and the targeted 15 mm polyp, extracted from CTC. In total, there were 370 surface structures evenly distributed over the colon surface that were used for registration. The global minimum of $\sigma$ is correctly located at the surface point cloud extracted from the 15 mm polyp ($\sigma_\mathrm{min}$ = 0.65).  

\section{Discussion}
Potentially, 3D reconstructions of polyps observed at OC could be matched with corresponding polyps reported at prior CTC. This would be useful for the colonoscopist, in order to assess whether the polyp he or she is seeing during the colonoscopic procedure is the same one as reported by the radiologist at prior CTC. Because it is known that CTC can miss polyps, by using automatically matched polyps the colonoscopist could more confidently determine whether an observed polyp is one previously reported or whether it is new (i.e. false-negative CTC). 

This phantom study illustrates how automatic polyp matching OC and CTC could be achieved, thus facilitating localization and verification of polyps detected at CTC. Subsequently, this will facilitate biopsy of true-positive pathology and exclusion of false-positive CTC findings. Consequently, it will also likely reduce the occurrence of false-negative or missed polyps by colonoscopy, when these have been seen previously at CTC. It might also help reduce healthcare costs, patient inconvenience and discomfort. This work assumes that bowel motion is relatively constrained (rigid) at the location of the polyp and the majority of perspective shift comes from the movement of the colonoscopic camera, which is likely to be true for short periods of time during colonoscopy. Potential incorporation of EM tracking data, or a way of automatically measuring the inserted length of the colonoscope, could provide a means of distinguishing between other potentially matching OC polyp locations with CTC.

\section{Acknowledgments} 
The authors gratefully acknowledge financial support for this work from the NIHR Program Grants for Applied Research: ``Imaging diagnosis of colorectal cancer: Interventions for efficient and acceptable diagnosis in symptomatic and screening populations'' (Grant No. RP-PG-0407-10338) and the EPSRC-CRUK Comprehensive Cancer Imaging Centre of UCL and KCL (Grant No. C1519AO).



\bibliographystyle{chicago} 
\bibliography{HRoth_spie2014_optical-to-ctc_submitted}

\end{document}